\documentclass[12pt]{article}
\usepackage{fullpage}
\usepackage{graphicx}
\usepackage{amsmath}
\usepackage{listings}
\usepackage{hyperref}
\usepackage{url}
\usepackage{subfig}
\usepackage[utf8]{inputenc}

\usepackage{algorithm}
\usepackage{algpseudocode}

\usepackage{setspace}
\title{Making forecasting self-learning and adaptive -- Pilot forecasting rack}
\author{Shaun D'Souza, Dheeraj Shah, Amareshwar Allati, Parikshit Soni \\ TCS}
\date{}

\begin{document}
\maketitle

\begin{abstract}

Retail sales and price projections are typically based on time series forecasting.  For some product categories, the accuracy of demand forecasts achieved is low, negatively impacting inventory, transport, and replenishment planning. This paper presents our findings based on a proactive pilot exercise to explore ways to help retailers to improve forecast accuracy for such product categories. 

We evaluated opportunities for algorithmic interventions to improve forecast accuracy based on a sample product category, Knitwear.  The Knitwear product category has a current demand forecast accuracy from non-AI models in the range of 60\%. We explored how to improve the forecast accuracy using a rack approach. To generate forecasts, our decision model dynamically selects the best algorithm from an algorithm rack based on performance for a given state and context. Outcomes from our AI/ML forecasting model built using advanced feature engineering show an increase in the accuracy of demand forecast for Knitwear product category by 20\%, taking the overall accuracy to 80\%. Because our rack comprises algorithms that cater to a range of customer data sets, the forecasting model can be easily tailored for specific customer contexts.

\end{abstract}

\section{Why do existing time series and supervised approaches to forecasting fall short?}

Currently, a variety of statistical models are used for time-series and supervised approaches to forecasting, namely Lewandowski and AVS Graves. 

\begin{itemize}
\item Lewandowski assumes that business is in a state of flux and changes at an inconstant rate. It considers features such as seasonality, trend, and demand and generates a forecast by evaluating actual sales data and analyzing patterns in the demand history. For establishing patterns, the elements of demand history considered are dynamic mean and seasonality, events, and external factors or lifecycles.
\item AVS Graves focuses on slow moving and intermittent (randomly distributed data with a high zero content) demand patterns. It uses a smoothing parameter that is automatically adjusted based on percentage error in each period. The adjustment occurs when the criterion for an update is satisfied or when demand is non-zero. It produces a flat line forecast which is only updated when demand occurs. The algorithm can model cyclical demand patterns. 
\end{itemize}

\begin{figure}[!h]
    \centering
	\includegraphics[width=0.9\columnwidth]{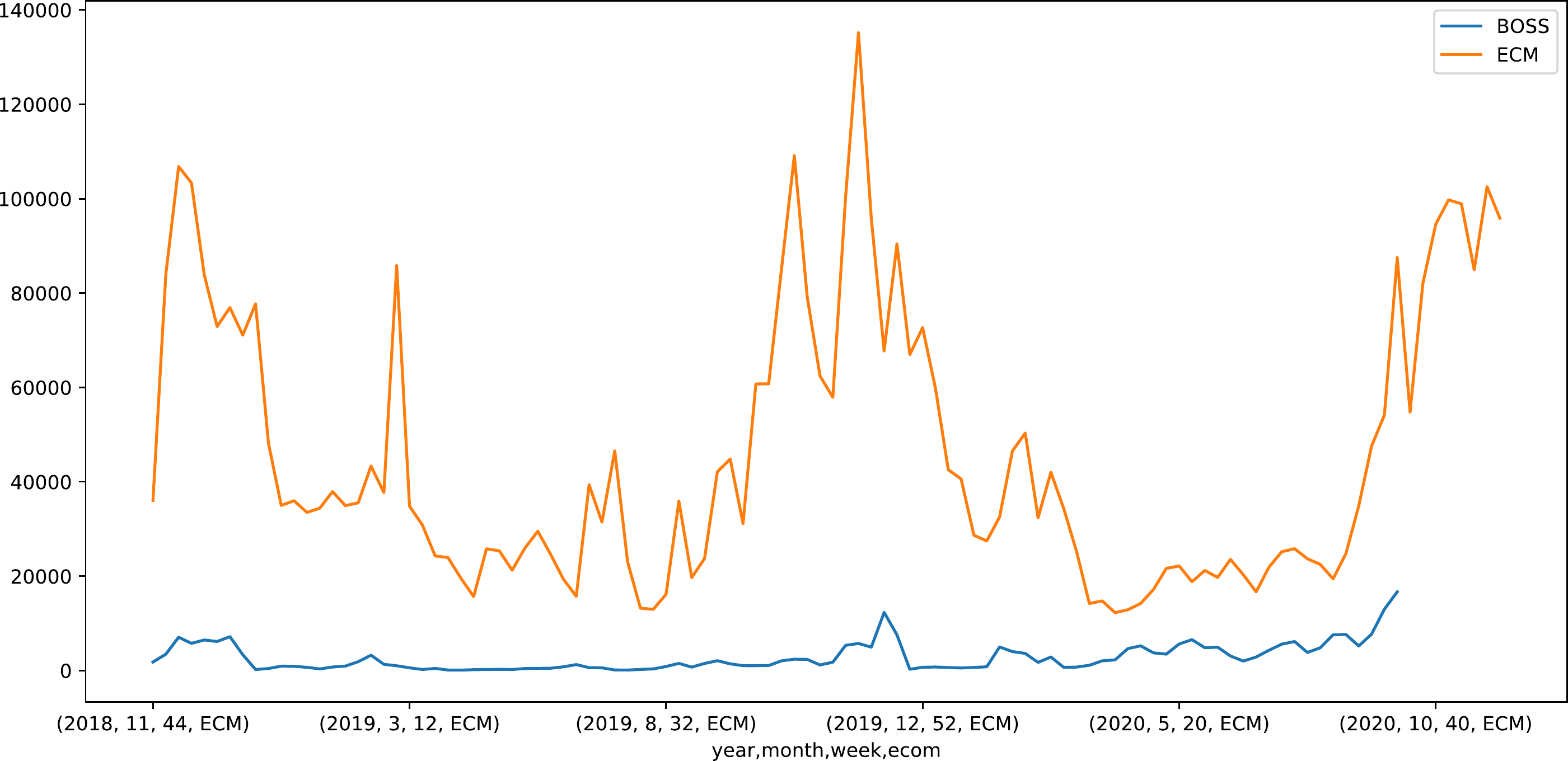}
	\caption{Aggregate sales Ecommerce/Brick and mortar stores}
	\label{fig:ecom-boss}
\end{figure}

Figure~\ref{fig:ecom-boss} shows a plot of the aggregated sales for e-commerce and brick and mortar stores in a unidimensional time series.

While these models provide adequate accuracy, they do not account for changing patterns in sales of items based on store, price, lag t - 1, t - 2, weather and other demographics. They offer little scope for customization, do not cater to specific pivots in the data, require manual intervention to manage exceptions, require parameters to be tweaked manually to incorporate additional influencers, and do not consider factors such as cannibalization, substitution, and affinity. The forecast accuracy is dependent on the experience of the user.

\section{Bridging the Performance Gaps in Demand Forecasting}

We developed a forecasting model that overcomes key limitations such as unaccounted key influencing parameters and absence of AI/ML based algorithmic model for demand prediction.  We created an algorithmic rack comprising multiple custom AI/ML based models for demand prediction and designed a decision model for dynamic selection of best-fit algorithm for given state and context. 

\begin{figure}[!h]
    \centering
	\includegraphics[width=0.5\columnwidth]{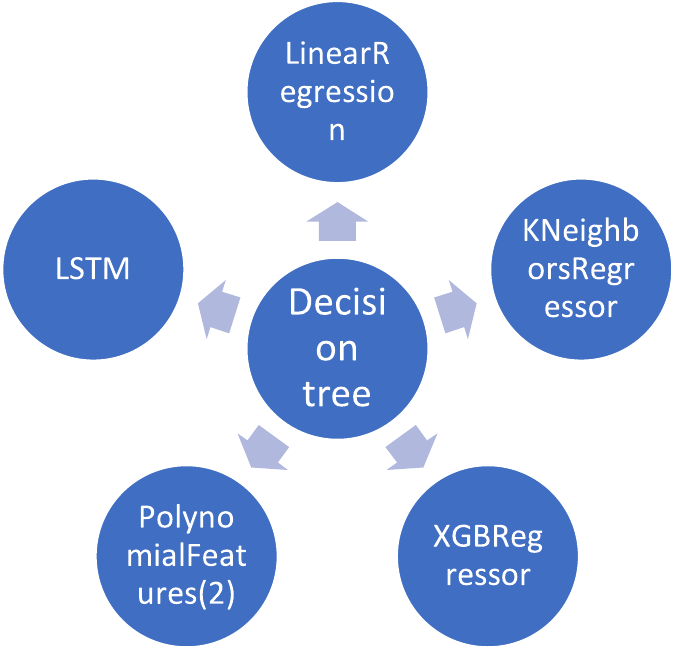}
	\caption{Algorithmic Rack}
	\label{fig:model-selector}
\end{figure} 

\textit{Vendor Demand + Adaptive Forecasting = Improved Forecast Accuracy for Knitwear category.}

\begin{figure}
    \centering
	\includegraphics[width=0.9\columnwidth]{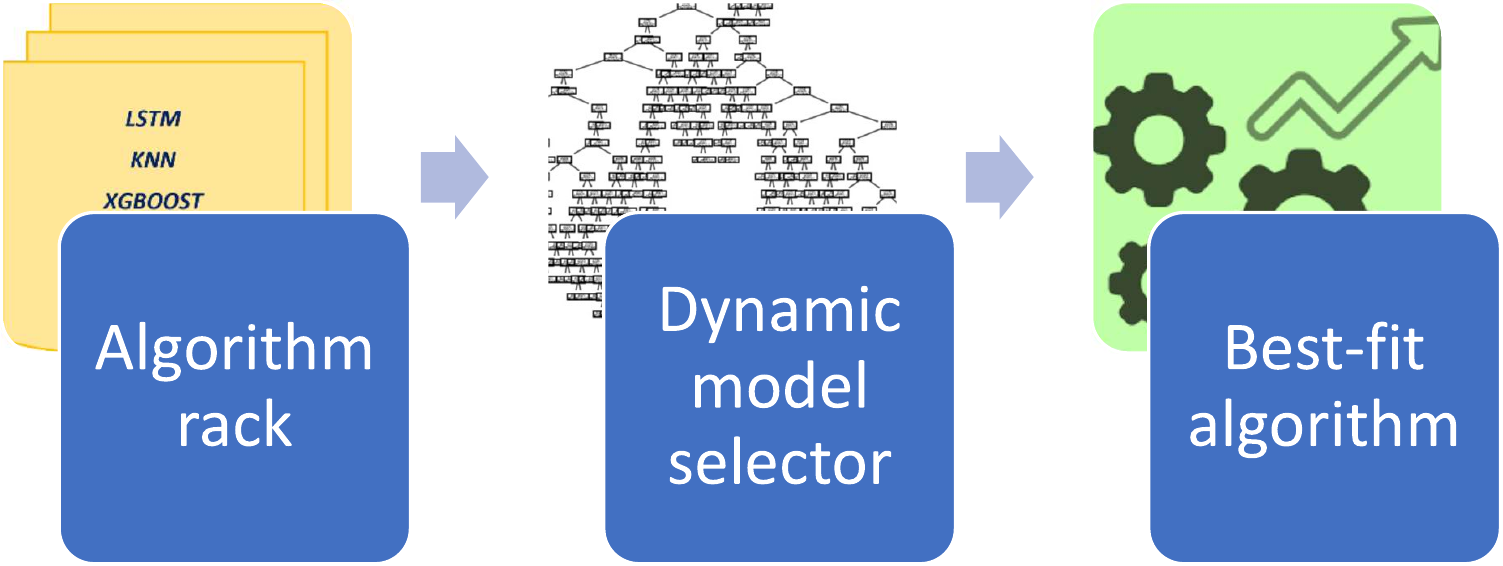}
	\caption{Adaptive Forecasting dynamic model selector}
	\label{fig:best-fit}
\end{figure} 

\section{Designing the Algorithmic Rack}

\nocite{supervised}

We modeled the seasonality's for the data using the Month, Week of year, Average Price, and Lag Sales t - 1 and t - 2. These parameters provide a baseline for our forecasting model. Pivots in sales data for Knitwear are below - Table \ref{table:knitwear}.

\begin{table}[!h]
\texttt{
\begin{tabular}{l l l l l l l} 
\hline
GAN &	Year &	Month &	WeekNo &	PromoAvailable & RangeID &	ItemID \\
\hline
22403673 &	2018 &	1 &	0 &	NO &	T38112 &	T3812A \\
22403673 &	2018 &	1 &	0 &	NO &	T38112 & 	T3812A \\
22403673 &	2018 &	1 &	0 &	NO &	T38112 &	T3812A \\
22403673 &	2018 &	1 &	0 &	YES &	T38112 &	T3812A \\
22403673 &	2018 &	1 &	1 &	NO &	T38112 &	T3812A \\
 & & & & & & \\
\end{tabular}
}

\texttt{
\begin{tabular}{l l l l l l l} 
\hline
FitID &	ListingInd &	Divison &	AvgSellPrice &	OriginalPrice &	MinTemp &	MaxTemp \\
\hline
F15 &	D &	Central &	36 &	75 &	0.9 &	4 \\
F15 &	D &	North &	36.5 &	75 &	1.5 &	4.227273 \\
F15 &	D &	South &	36 &	75 &	2.078125 &	5.234375 \\
F15 &	D &	South &	58.25 &	75 &	2 &	5.125 \\
F15 &	D &	Central &	31.42125 &	75 &	4.125 &	6 \\
 & & & & & & \\
\end{tabular}
}

\texttt{
\begin{tabular}{l l l l l l} 
\hline
HrsSunShine &	HrsRainfall &	HrsSnowFall &	HrsPercipitation &	SalesQty \\
\hline
1.116667 &	0.733333 &	0.633333 &	1.366667 &	37 \\
1.295455 &	0.431818 &	0.954545 &	1.386364 &	21 \\
1.421875 &	0.765625 &	1.15625 &	1.96875 &	22 \\
0.75 &	0.125 &	0.5 &	0.625 &	2 \\
0.375 &	0 &	0 &	0 &	10 \\
\end{tabular}
}

\caption{Sales data for Knitwear}
\label{table:knitwear}
\end{table}

Over the last few decades, with the rise of the web, social media, and trends such as viral media and influencers, demand trends have slowly and inexorably become more volatile and less suited to traditional demand forecasting techniques, especially in the case of product categories such as fashion, apparel, footwear, and technology. Based on our unique and exhaustive customer dataset, we were able to successfully synthesize a dump of the historical data over a number of features which we determined were key to a more accurate demand forecasting system. Some of the features considered for our adaptive forecasting use case are shown in Listing 1.

\textbf{Listing 1: Feature consideration to generate superior forecasts}

\textbf{Available and can be used:} 
\begin{enumerate}
\item Sales quantity
\item Sales price
\item Listing period
\item Seasonality
\item Weather condition
\item Store status (live/closed)
\item Out of stock days count
\item Product attributes
\end{enumerate}

\textbf{Available but not usable:}
\begin{enumerate}
\item Customer sentiments (social media scrapes)
\item Product affinity
\item Campaign
\item Store footfall
\item New product launches
\item Events (social, local/national)
\item Clickstream tracking
\item Competitor promotion
\item Product review ratings
\item Competitor campaign
\item Competitor pricing		
\end{enumerate}
	
\textbf{Not available:}
\begin{enumerate}
\item Customer return quantity
\item Promotions
\end{enumerate}

\begin{figure}[!h]
    \centering
	\includegraphics[width=.5\textwidth]{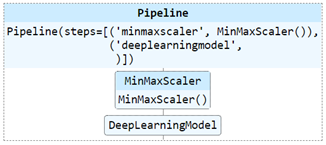}
	\caption{Deep Learning MinMaxScaler Pipeline}
	\label{fig:minmax}
\end{figure}

\begin{figure}[!h]
    \centering
	\includegraphics[width=.4\textwidth]{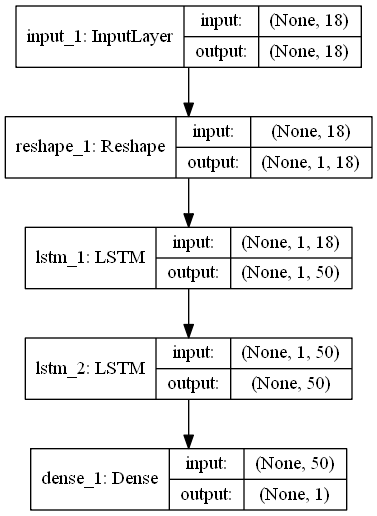}
	\caption{Keras LSTM DeepLearningModel}
	\label{fig:lstm-model}
\end{figure}

The Deep Learning LSTM model \cite{keras} is depicted in Figure~\ref{fig:minmax}, Figure~\ref{fig:lstm-model}.

Figure~\ref{fig:rmse},~\ref{fig:r2} below show the RMSE and R2 errors for each of the trained models.
 
\begin{figure}[!h]
\begin{minipage}{.5\textwidth}
    \centering
	\includegraphics[width=.9\textwidth]{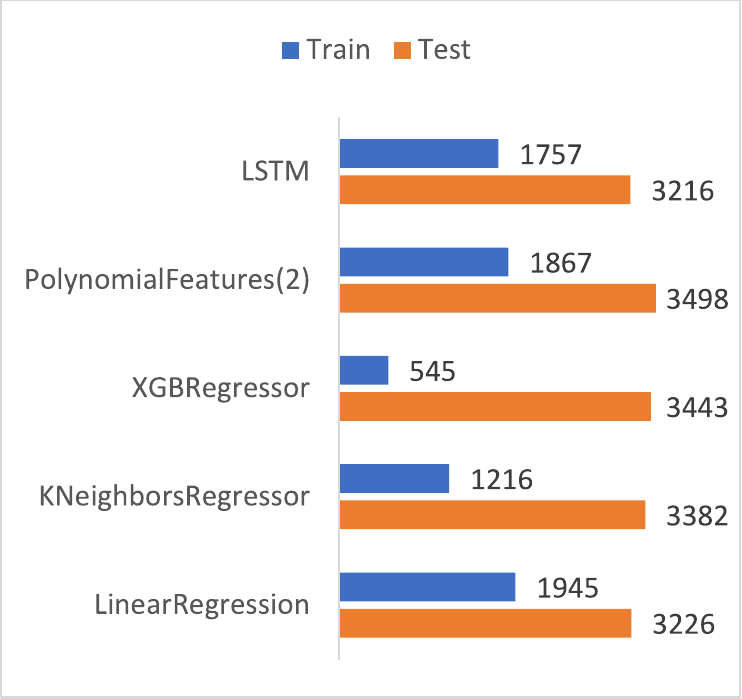}
	\caption{RMSE error}
	\label{fig:rmse}
%\end{figure}
\end{minipage}
\begin{minipage}{0.5\textwidth}
%\begin{figure}[!h]
    \centering
	\includegraphics[width=.9\textwidth]{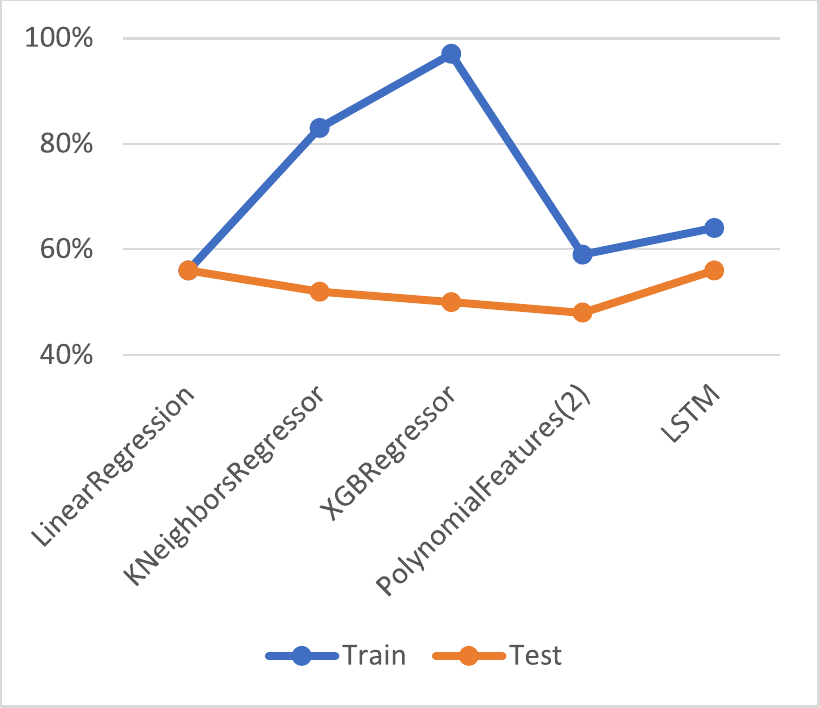}
	\caption{R2 error}
	\label{fig:r2}
\end{minipage}
\end{figure}

The data shows that the XGBoost Regressor produces the smallest RMSE and high R2 in the train and test sets, making it the most suitable for our regression modeling. This is followed by the LSTM and K-nearest neighbors models.

We tested and evaluated the above findings  for channel-wise sales based on eCommerce and brick and mortar stores. These include:

\begin{itemize}
\item Article -- Store -- Country
\item Article -- Ecommerce -- Country
\item Article -- All channels -- Country 
\end{itemize}

\section{Implementing Adaptive Forecasting with a dynamic model selector}

We implemented a Decision Tree based dynamic model selector to determine the most suitable forecasting algorithm. Historical data is used in the determination of the choice of ML model. These decisions are made in advance to detect changes in consumption behaviors. The selector chooses the best model from the historical data \cite{decision}. It assumes that the past is a guidance into the future sales forecasts and can forecast any major demand shifts in the inventories arising out of seasonality's such as upcoming festivals.
 
A training dataset containing information related to Article ID and Week of sales is used along with the actual and predicted sales numbers. These are consolidated for each of our 5 regression models - Table \ref{table:regression}.

\begin{table}
\begin{center}
\texttt{
\begin{tabular}{|l|l|} 
\hline
Linear & sklearn.linear\_model.LinearRegression \\
K-nearest Neighbours & sklearn.neighbors.KNeighborsRegressor \\
XG Boost \cite{xgboost} & xgboost.XGBRegressor \\
Polynomial (squared) & sklearn.preprocessing.PolynomialFeatures \\
Deep Learning LSTM model & keras.layers.LSTM \\
\hline
\end{tabular}
}
\end{center}
\caption{ML Regression models}
\label{table:regression}
\end{table}

\nocite{gluonts}

We measure the Absolute Percentage Error (APE) for the pivots in our dataset using:

$$ APE = \frac {| y_{true} - y_{pred} |}{y_{true}} $$

\begin{figure}[H]
    \centering
%	\hspace*{-2cm}
    \subfloat[]{{\includegraphics[height=0.2\textheight]{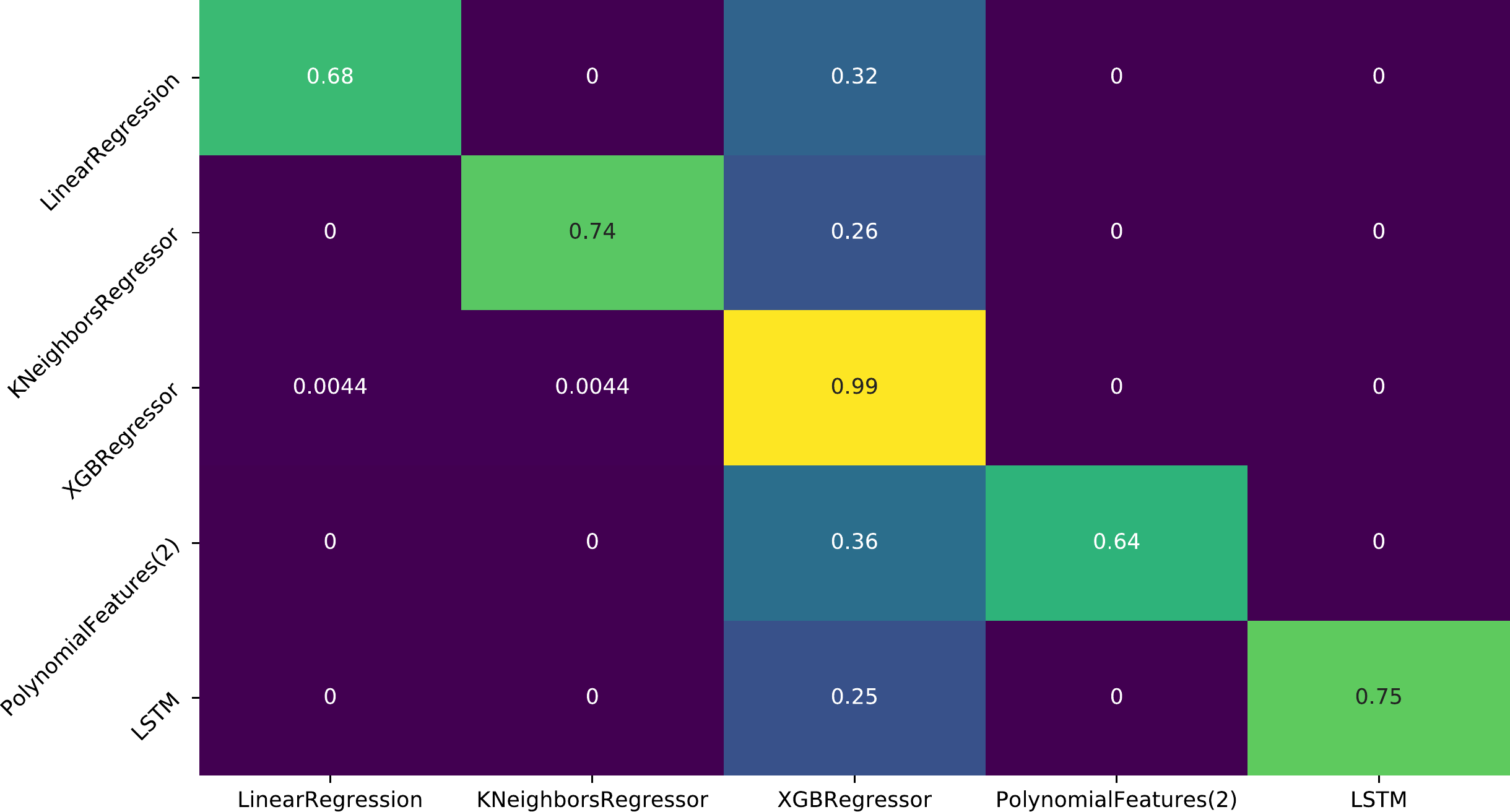}}}
	\vspace{10pt}
    \subfloat[]{{\includegraphics[height=0.2\textheight]{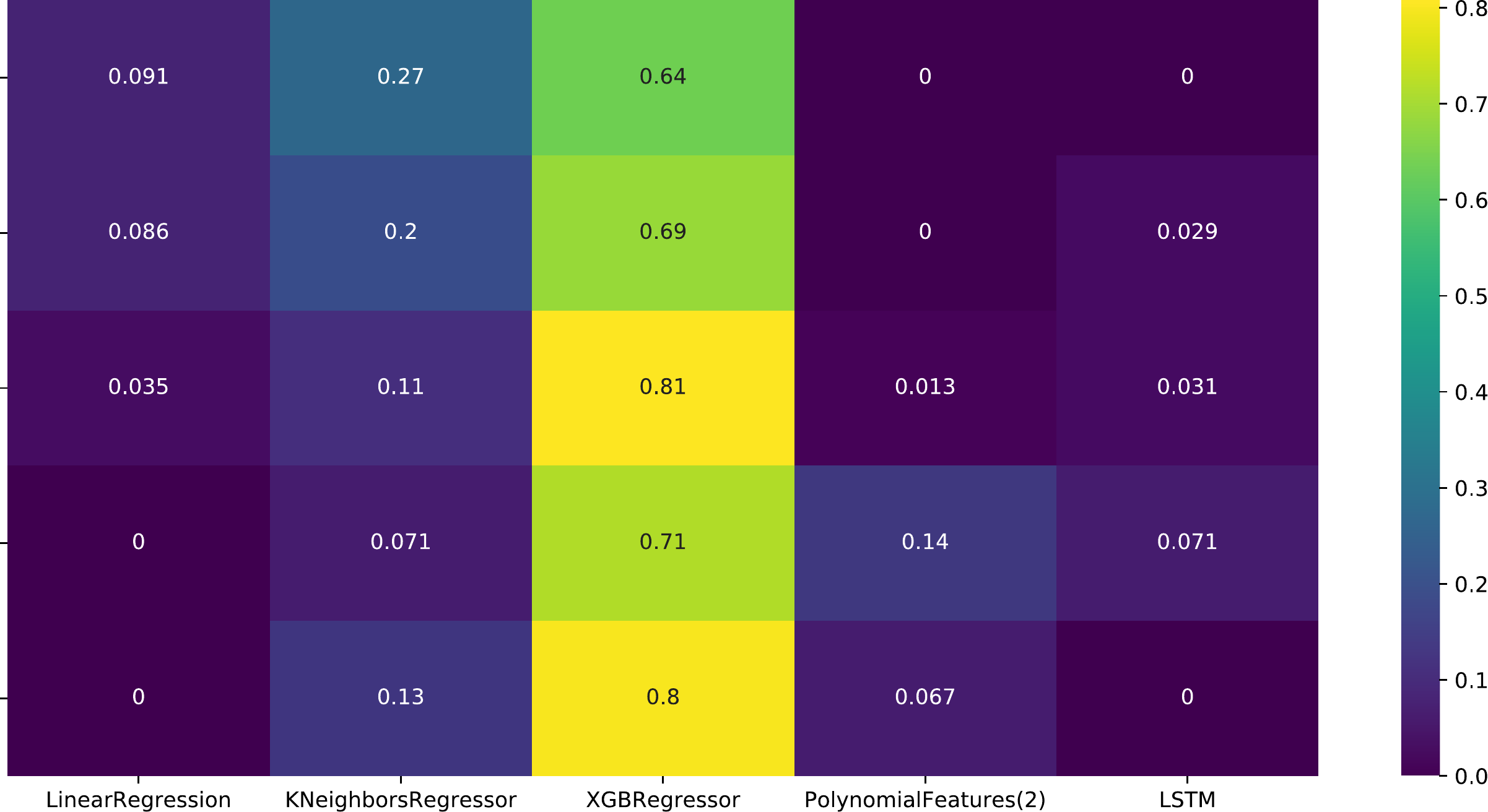}}}
	\newline
    \subfloat[]{{\includegraphics[width=0.8\textwidth]{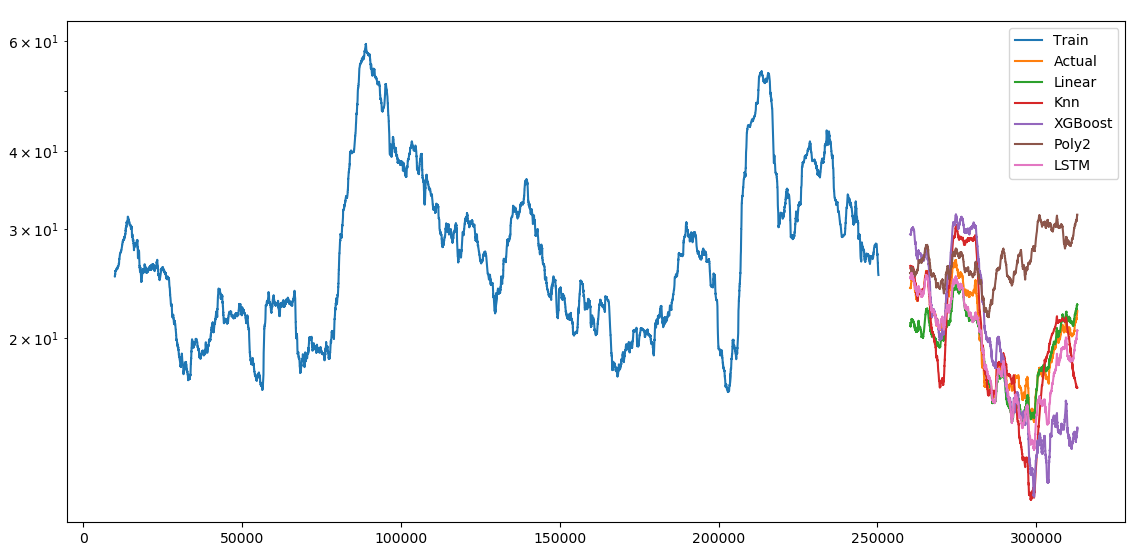}}}
	\caption{Confusion matrix (a) Train (b) Test (c) Average sales forecast actual vs. predicted}
	\label{fig:avg-sales}
\end{figure} 

\begin{algorithm}[!h]
\caption{Model selection using APE}\label{alg:selection-ape}
\begin{algorithmic}[1]

\Require $y_{true} \geq 0$
\Ensure $y_{pred} \geq 0$
\State $error \gets \Call{pandas.DataFrame}{{}}$
\ForAll{$pred \in \{linear, knn, xgboost, poly2, lstm\}$} 
\Comment{Calculate error}
\State $error[pred] \gets \Call{APE}{y_{true}, y_{pred}}$
\EndFor

\State $X \gets features$
\State $y \gets \Call{np.argmin}{error}$ 
\Comment{Model selector}
\State \textbf{output} [1, 2, 4, 1, \dots]

\State $model\_selector \gets \Call{sklearn.tree.DecisionTreeClassifier().fit}{X, y}$ 
\Comment{Train}
\State \Return model\_selector
\end{algorithmic}
\end{algorithm}

A categorical dataset that uses numpy.argmin selects the model with the least error - Algorithm~\ref{alg:selection-ape}. This is input to the dynamic selector model. We implemented a Decision tree to model the characteristics of the selector. This is fit of the classification dataset to determine the most suitable forecasting model.

\section{Conclusion}

Figure~\ref{fig:avg-sales} (c) shows the average forecasted sales for each of the trained ML models. It highlights the actual vs. predicted trend for our dataset with $\sim$ 3 million data points. 

The Deep Learning LSTM and Linear Regression models were found to train efficiently for larger datasets such as Sales data containing both GAN and Store information. The XGBoost and K-nearest neighbours models were constrained because the training set consisted of \textgreater 1 million records.

\bibliography{refs}
\bibliographystyle{plain}

\end{document}